\documentclass[10pt,letterpaper,twocolumn]{article}
\usepackage[latin9]{inputenc}
\usepackage{array}
\usepackage{multirow}
\usepackage{amsmath}
\usepackage{amssymb}
\usepackage[unicode=true,
 bookmarks=false,
 breaklinks=true,pdfborder={0 0 1},backref=section,colorlinks=false]
 {hyperref}

\makeatletter

\pdfpageheight\paperheight
\pdfpagewidth\paperwidth

\providecommand{\tabularnewline}{\\}

\usepackage{cvpr}
\usepackage{times}
\usepackage{epsfig}
\usepackage{graphicx}

\usepackage{import}
\usepackage{breqn}

\usepackage{multirow}
\cvprfinalcopy % *** Uncomment this line for the final submission

\def\cvprPaperID{7210} % *** Enter the CVPR Paper ID here

\makeatother

\begin{document}
%%%%%%%%% TITLE

\title{FOSS: Multi-Person Age Estimation with Focusing on Objects and Still
Seeing Surroundings}
\author{Masakazu Yoshimura\\
 The University of Tokyo\\
\texttt{\small{}yoshimura4812@g.ecc.u-tokyo.ac.jp}\and{\small{} }Satoshi
Ogata\\
 BTC Corporation, Japan{\small{}}\\
{\small{} }\texttt{\small{}satoshi.ogata@bigtreetc.com}}
\maketitle
\begin{abstract}
Age estimation from images can be used in many practical scenes. Most
of the previous works targeted on the estimation from images in which
only one face exists. Also, most of the open datasets for age estimation
contain images like that. However, in some situations, age estimation
in the wild and for multi-person is needed. Usually, such situations
were solved by two separate models; one is a face detector model which
crops facial regions and the other is an age estimation model which
estimates from cropped images. In this work, we propose a method that
can detect and estimate the age of multi-person with a single model
which estimates age with focusing on faces and still seeing surroundings.
Also, we propose a training method which enables the model to estimate
multi-person well despite trained with images in which only one face
is photographed. In the experiments, we evaluated our proposed method
compared with the traditional approach using two separate models.
As the result, the accuracy could be enhanced with our proposed method.
We also adapted our proposed model to commonly used single person
photographed age estimation datasets and it is proved that our method
is also effective to those images and outperforms the state of the
art accuracy.
\end{abstract}
%%%%%%%%% BODY TEXT

\section{Introduction}

\footnotetext{This work was completed while the first author was an intern at BTC Corporation.
\\© 2020 M. Yoshimura and  S. Ogata.}

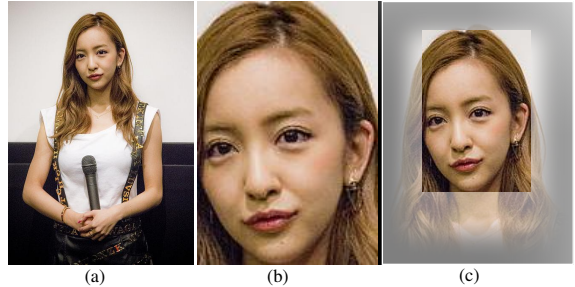
\begin{figure}
\centering

\def\svgwidth{0.9\columnwidth}

\scriptsize%% Creator: Inkscape 1.0.1 (3bc2e813f5, 2020-09-07), www.inkscape.org
%% PDF/EPS/PS + LaTeX output extension by Johan Engelen, 2010
%% Accompanies image file 'concept.pdf' (pdf, eps, ps)
%%
%% To include the image in your LaTeX document, write
%%   \input{<filename>.pdf_tex}
%%  instead of
%%   \includegraphics{<filename>.pdf}
%% To scale the image, write
%%   \def\svgwidth{<desired width>}
%%   \input{<filename>.pdf_tex}
%%  instead of
%%   \includegraphics[width=<desired width>]{<filename>.pdf}
%%
%% Images with a different path to the parent latex file can
%% be accessed with the `import' package (which may need to be
%% installed) using
%%   \usepackage{import}
%% in the preamble, and then including the image with
%%   \import{<path to file>}{<filename>.pdf_tex}
%% Alternatively, one can specify
%%   \graphicspath{{<path to file>/}}
%% 
%% For more information, please see info/svg-inkscape on CTAN:
%%   http://tug.ctan.org/tex-archive/info/svg-inkscape
%%
\begingroup%
  \makeatletter%
  \providecommand\color[2][]{%
    \errmessage{(Inkscape) Color is used for the text in Inkscape, but the package 'color.sty' is not loaded}%
    \renewcommand\color[2][]{}%
  }%
  \providecommand\transparent[1]{%
    \errmessage{(Inkscape) Transparency is used (non-zero) for the text in Inkscape, but the package 'transparent.sty' is not loaded}%
    \renewcommand\transparent[1]{}%
  }%
  \providecommand\rotatebox[2]{#2}%
  \newcommand*\fsize{\dimexpr\f@size pt\relax}%
  \newcommand*\lineheight[1]{\fontsize{\fsize}{#1\fsize}\selectfont}%
  \ifx\svgwidth\undefined%
    \setlength{\unitlength}{403.61997986bp}%
    \ifx\svgscale\undefined%
      \relax%
    \else%
      \setlength{\unitlength}{\unitlength * \real{\svgscale}}%
    \fi%
  \else%
    \setlength{\unitlength}{\svgwidth}%
  \fi%
  \global\let\svgwidth\undefined%
  \global\let\svgscale\undefined%
  \makeatother%
  \begin{picture}(1,0.50920665)%
    \lineheight{1}%
    \setlength\tabcolsep{0pt}%
    \put(0,0){\includegraphics[width=\unitlength,page=1]{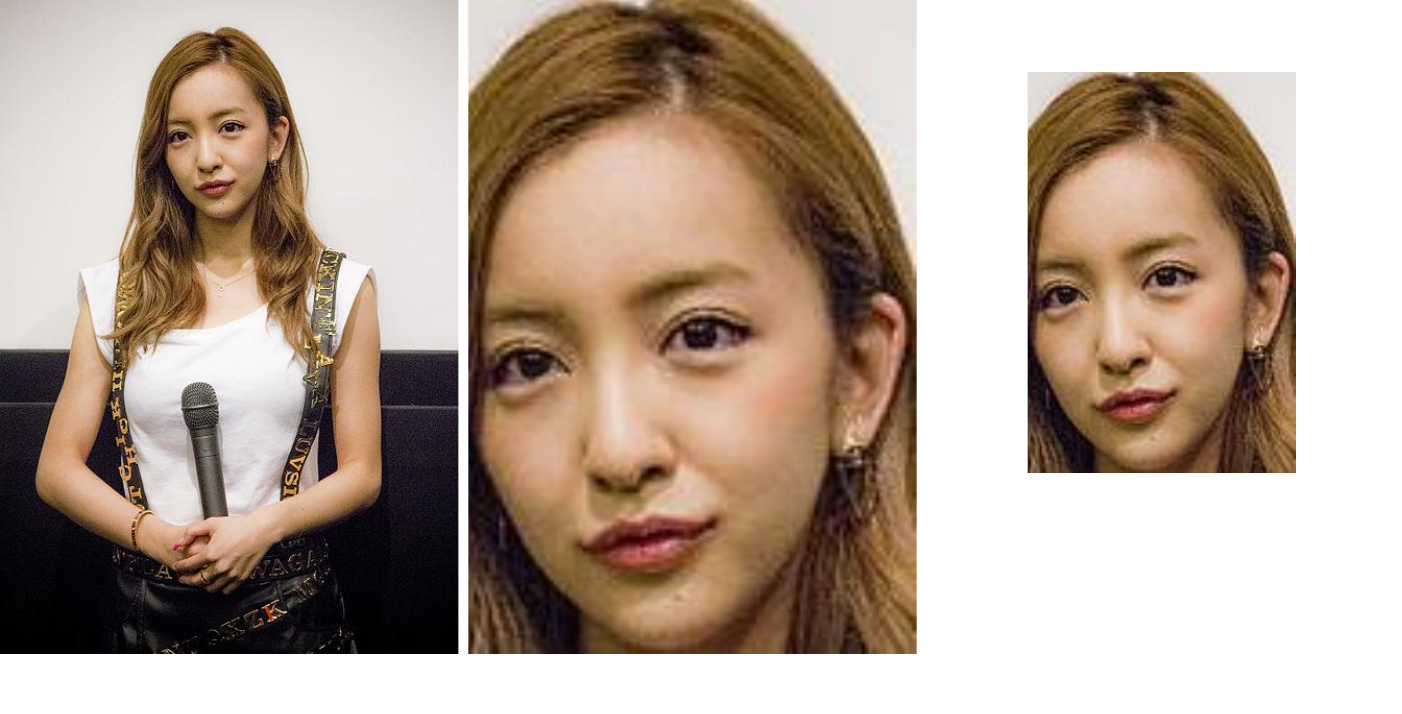}}%
    \put(0.1343079,0.01109155){\color[rgb]{0,0,0}\makebox(0,0)[lt]{\lineheight{1.25}\smash{\begin{tabular}[t]{l}(a)\end{tabular}}}}%
    \put(0.45772611,0.01071952){\color[rgb]{0,0,0}\makebox(0,0)[lt]{\lineheight{1.25}\smash{\begin{tabular}[t]{l}(b)\end{tabular}}}}%
    \put(0.79770443,0.01316065){\color[rgb]{0,0,0}\makebox(0,0)[lt]{\lineheight{1.25}\smash{\begin{tabular}[t]{l}(c)\end{tabular}}}}%
  \end{picture}%
\endgroup%

\caption{\label{fig:The-visualization-of}The visualization of the concept.
(a) has the richest information, meanwhile (b) is easy to extract
facial information. (c) is focusing on the face but still seeing surroundings.}
\end{figure}

Age estimation from images is an important research task because it
can be used in many practical scenes like surveillance, human-robot
interaction, recommendation, customer age group survey in a shop,
and so on. It was also used in some other research tasks as a part
of the methods. For instance, it was used to train Generative Adversarial
Networks (GAN) \cite{goodfellow2014generative} for facial age editing
\cite{yao2020high} and used to support facial recognition network
\cite{zheng2017age}. It is still a challenging task mostly due to
the difference between apparent age and real age caused by genetic
difference, makeup, angle, and facial expression.

In recent years, convolutional neural networks (CNN) based methods
achieved great success in age estimation \cite{zhang2019c3ae,li2019bridgenet,cao2019rank,pan2018mean,gustafsson2019dctd,berg2020deep}.
Although their methods are varied from each other, most of their research
target is to estimate from images like pictures for identification
certificates in which only one face exists. However, in some situations
like human-robot interaction or automatic customer age group surveys
in a shop, age estimation in the wild and for multi-person is needed.
Our goal of this paper is to estimate the age of multi-person in the
wild accurately. Such situations were usually solved by using an additional
face detector which crops facial regions and creates images to feed
to the age estimator \cite{eidinger2014age}. Although it might be
a bit different from multi-person age estimation, \cite{Rothe-ICCVW-2015}
cropped a facial region from the huge number of wild images crawled
from the internet and used them to pre-train the age estimator model.
Also further tight cropping from already cropped facial datasets is
commonly used as a preprocessing to align the images \cite{zhang2019c3ae,pan2018mean,li2019bridgenet,chen2017using}.
Here is our idea come from. Is it optimal in terms of accuracy and
efficiency to use two separate models of a detector and an age estimator
and to estimate age from tightly cropped images? Though it is reasonable
that alignment with tight cropping makes easy to estimate and increases
the accuracy since it puts facial parts always at the almost same
place and puts more focus on the face, it loses some information of
surrounding pixels. The information of the surrounding pixels might
be useful to figure out whether the texture on the face is derived
from aging, characteristic of the camera, or the environment. From
these perspectives, we propose a novel model that focuses on the face
the same with previous methods but, at the same time, still considers
surroundings differently from previous works.

To train a multi-person age estimation model, datasets which include
multi-face photographed images and corresponding annotation of the
facial locations and ages are desired, but those open datasets are
rare. To the best of our knowledge only Images of Groups (IoG) dataset
\cite{gallagher2009understanding} is such a dataset. However, the
age annotation was defined with 7 age groups and it is too rough to
make the model estimate coarsely. On the other hand, many single face
photographed datasets were proposed with the rich amount and per year
level annotation \cite{zhang2017quantifying,Rothe-ICCVW-2015,zhang2017age}.
These rich and coarse datasets could be created thanks to automatic
crawling from the Internet. Even if sets of multiple faces and their
age information exist on the Internet, to correspond each age information
to each face is difficult. So multi-person photographed dataset is
difficult to create. Based on the above, we also propose a training
strategy for the multi-person age estimation model. In this method,
a dataset for face detection and single face photographed age estimation
datasets are used to train the model.

In short, our goal is to estimate the age of multi-person accurately
and the contributions of our paper are as follows:
\begin{itemize}
\item A novel model that can estimate the age of multi-person different
from previous methods. Experiments showed that it is more accurate
than using two separate models of face detection and age estimation
despite having a little fewer parameters.
\item The conceptual architecture of focusing on the face still somewhat
considering surroundings is effective even to the images only one
face is photographed and it outperformed the state-of-the-art on the
two open datasets; IMDB-WIKI \cite{Rothe-ICCVW-2015} and UTKFace
\cite{zhang2017age}.
\item A training strategy for a multi-person age estimation model with limited
datasets is proposed. It enables to train well with widely available
datasets; a face detection dataset and single face photographed datasets.
\end{itemize}

\section{Related Works}

\subsection{Face Detection}

Although most of the previous age estimation methods aimed to estimate
from single face photographed images, they usually utilized face detection
methods to crop tightly and align the images as a preprocessing. For
example, BridgeNet \cite{li2019bridgenet} utilized a face detection
method which could run 99 fps with a GPU \cite{guo2013joint}. Mean
variance loss \cite{pan2018mean} utilized a detection method \cite{wu2017funnel}
which connected multiple cascade classifiers as a funnel-like structure
and can process with 20 fps without using GPUs. Thus, age estimation
methods for single face photographed images tend to utilize small
and fast face detection method. This may be because the usual datasets
for age estimation photograph only one face (although some images
contain multiple faces mistakenly) and detection from them is easy.

On the other hand, deep CNN based face detection methods were proposed
and beat previous methods in several years. AInnoFace \cite{zhang2019accurate}
and RetinaFace \cite{deng2019retinaface} were based on RetinaNet
\cite{lin2017focal} and obtained high accuracy on WiderFace dataset
\cite{yang2016wider}, a challenging dataset containing for face detection.
In the more recent years anchor free detection methods were proposed
in the field of common objects detection \cite{zhou2019objects,law2018cornernet}.
CenterNet \cite{zhou2019objects} is a keypoint based detection method
and surpass RetinaNet in terms of both accuracy and speed on COCO
dataset \cite{lin2014microsoft}, a dataset for common objects detection.
So, our detection part is based on CenterNet. Although \cite{deng2019retinaface}
reported that estimating the bounding boxes with its facial landmarks
enhanced the accuracy, we don't utilize facial landmarks because the
gain was small and many of the age estimation datasets don't contain
annotations of facial landmarks. 

\subsection{Age Estimation}

As far as we know, most of the previous works were targeted on estimating
age from single person photographed images. These methods can be categorized
into four groups by the way of obtaining estimation. One is a method
that estimates age as a regression problem \cite{guo2012study,guo2009human}.
In other words, it tries to output the true value of the age. However,
it was reported that regression methods tend to overfit the training
data because apparent age and real age have a gap, and the model fit
to the gaps in the training data \cite{chang2011ordinal}. The second
one is a method which estimates age as a classification problem \cite{rodriguez2017age,agustsson2017anchored,Rothe-ICCVW-2015}.
In the experiment of DEX \cite{Rothe-ICCVW-2015}, it is reported
that if the amount of the training data is enough, classifying per
year (classifying into 101 labels) is more accurate than regression.
Also, it reported that to make the final output as the weighted average
of the probability of the all age classes instead of outputting the
most probable age class enhanced the accuracy. The third one outputs
the probability of the ages almost the same as the classification
method but treats the ordinal relationship between the age classes.
C3AE \cite{zhang2019c3ae} proposed a two-points representation way
of outputs. Its age classes were rough and around per 10 ages but
it could estimate coarsely from the probabilities of two adjacent
classes. Also, Mean variance loss was proposed to cope with not only
the ordinal information but also the ambiguity of apparent ages by
estimating possible age distributions instead of exact ages \cite{pan2018mean}.
The last one is the ranking based method which classifies the ages
several times from roughly to coarsely like the concept of Decision
Forest \cite{chen2017using,li2019bridgenet}. BridgeNet connected
neighbor nodes of the tree structure and obtained robustness different
from the previous ranking based methods.

Judging by the accuracy and computational burden, we created the age
estimation part based on Mean variance loss \cite{pan2018mean}.

\subsection{Detection and Other Tasks in a Single Model}

In other fields, several methods dealt with the detection task and
subsequent other tasks in a single model. For example, several instance
segmentation approaches firstly estimated bounding boxes from the
feature obtained from its backbone network and then estimate segmentation
masks per bounding boxes using the same feature obtained from its
backbone network \cite{he2017mask,fang2019instaboost}. Mask-RCNN
\cite{he2017mask} demonstrated that it was also possible to estimate
human pose with the same pipeline of the instance segmentation approach.
Also, an anchor free and keypoint-based detection method of CenterNet
could estimate bounding boxes and human poses at one time \cite{zhou2019objects}.
In another field, several methods estimated the 3D pose of rigid objects
together with their bounding boxes at one time \cite{kehl2017ssd,9196779}.
In the similar field to age estimation, smile recognition and facial
attribute prediction were conducted at the same time with face detection
\cite{jang2019registration}.

Our proposed architecture is different from these previous methods
in terms of using two different backbones for the detection part and
the subsequent age estimation part. As described in the next section,
this is more computationally efficient for age estimation. 

\section{Proposed Method}

We propose a multi-person age estimation method that can detect faces
and estimate their ages with a single model different from previous
methods. In this section, we firstly present our model architecture
that aims not only to enable multi-person age estimation but also
to estimate more accurately by focusing on faces at the same time
somewhat seeing the surrounding environment. Then the loss function
to train the model is described. Finally, a training strategy is proposed
to train the multi-person age estimation model with limiting available
datasets: a face detection dataset and single face photographed datasets.
This strategy is needed because creating a multi-person photographed
age estimation dataset is difficult.

\subsection{Model Architecture}

\begin{figure*}
\centering

\def\svgwidth{1.9\columnwidth}

\scriptsize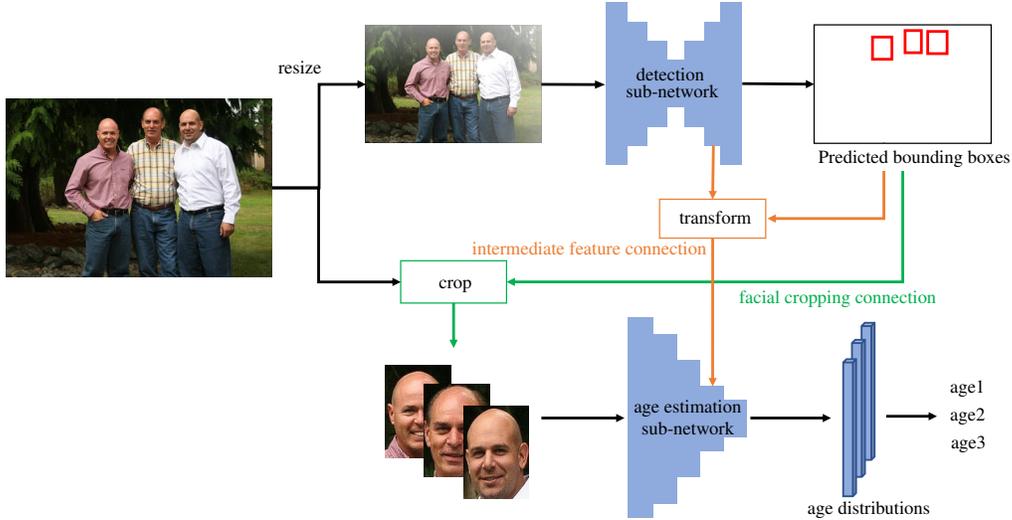

\caption{\label{fig:Overview-of-the}Overview of the proposed multi-person
age estimation}
\end{figure*}

An overview of our model for multi-person age estimation is shown
in Fig. \ref{fig:Overview-of-the}. It consists of a face detection
sub-network and an age estimation sub-network. Our motivations for
this architecture are as follows. If utilizing previous single person
age estimation methods on multi-person age estimation, we have to
utilize other face detection method to crop faces to feed to the age
estimation methods. It might be time-consuming to use two separate
models. Also, the accuracy might decrease when implemented in real
applications if the age estimator is overfitted to the alignment of
the training data. Most of the age estimation datasets contain largely
photographed faces but, in the multi-person scenario, faces are shown
with small sizes. Many detectors tend to react differently against
different sizes of objects due to the difference of the resolution,
default anchor shapes, or how many convolution layers the features
are passed through. The other motivation of the architecture is to
support age estimation by the surrounding pixels. There are two connections
between the detection sub-network and the age estimation sub-network.
One is a facial cropping connection which is aimed to align and focus
on the face to make the network easily compare different faces. The
other connection is the intermediate feature connection which gives
surrounding information to some extent to support age estimation.
the details are as follows.

\subsubsection*{Facial Cropping Connection}

The facial cropping connection crops facial regions corresponding
to the bounding boxes estimated with the detection sub-network. This
aims to align and focus on the faces similarly to previous works.
We crop with some margin from bounding box estimation to be robust
to the detection error. And the input images for age estimation are
cropped from high-resolution images to include detail information
like wrinkles. For the sake of speed and multi-process with GPU, always
$K$ most probable regions are cropped. 

\subsubsection*{Intermediate Feature Connection}

In the intermediate feature connection, the intermediate feature of
the detection sub-network is transformed and mixed with the intermediate
feature of the age estimation sub-network by concatenation. The transformation
corresponds to cropped regions of the facial cropping connection.
In other words, the corresponding regions of the intermediate feature
per face are affine transformed with bilinear interpolation. Although
the corresponding facial regions of the intermediate feature are extracted,
these transformed features contain the information outside the regions
to some extent. If $N$ layers with $3\times3$ convolutions are stacked,
the possible receptive field of the one point of the last feature
becomes $(2N+1)\times(2N+1)$ with the center pixels in the input
image more influential and the edge pixels less influential. So, by
concatenating the intermediate feature, to estimate age with the surrounding
pixel information is realized. As mentioned above, the surrounding
pixels might help to judge the texture on faces is whether derived
from aging, camera noise, or environments.

\subsubsection*{Separate Backbones}

Most of the previous works which dealt with detection and subsequent
tasks in a single model usually shared the backbone as described in
the section 2.3. Usually, it is memory efficient but, for age estimation,
it is not the case because facial detailed texture like wrinkles needs
to retain in the input image. For example, if we want to detect faces
whose sizes are around 1/10 of the image size and $150\times150$
size is needed to retain detailed texture, the input image to the
shared backbone should be $1500\times1500$ size. It needs heavy computation.
In this context, detection and age estimation backbones are separated.
To the detection backbone, resized images are input and, for the age
estimation backbone, cropped images from high-resolution ones are
input.

\subsection{Loss Function}

To train the model, we define multi-task loss,

\[
L=L_{det}+L_{age}+L_{gen}
\]
where $L_{det}$ is a loss for the detection part and $L_{age}$,
$L_{gen}$are losses for the age and gender estimation part. Gender
was also estimated with some datasets together with age in the experiments.
Because we utilized CenterNet \cite{zhou2019objects} as a detection
sub-network in the experiments, the $L_{det}$ is defined as,

\[
L_{det}=L_{reg}+\lambda_{size}L_{size}+\lambda_{off}L_{off}
\]
the same with the original method. $L_{reg}$ is a regression loss
for keypoint maps, $L_{size}$ is a loss for bounding boxes' size,
and $L_{off}$ is a loss for sub-pixel adjustment of bounding boxes.
$\lambda_{suffix}$ denotes a weighting factor for $suffix$ and it
is a hyperparameter. The loss for age estimation is defined as,

\begin{align*}
L_{age} & =\frac{1}{N_{a}}\sum_{b}^{B}\sum_{k}^{K}b_{iou}\left(\hat{d_{b,k}},\,d_{b}\right)b_{age}\left(d_{b,k^{\prime}}\right)\\
 & \times L_{age\_single}\left(\hat{a_{b,k}},\,a_{b,k^{\prime}}\right)
\end{align*}

where $B$ denotes batch size, $K$ is the maximum number of faces
per image, $\left\{ \hat{d_{b,k}},\,d_{b}\right\} ,\left\{ \hat{a_{b,k}},\,a_{b,k^{\prime}}\right\} $are
prediction and ground truth of bounding boxes and age probability
vectors. $b_{iou}\left(\hat{d_{b,k}},\,d_{b,}\right)$ is a boolan
function,

\[
b_{iou}\left(\hat{d_{b,k}},\,d_{b}\right)=\begin{cases}
1 & max\,IOU\,>\,th_{iou}\\
0 & else
\end{cases}
\]
in which, firstly the predicted bounding box is matched with all of
the ground truth bounding boxes and calculated their intersection
over union (IOU). Then, if the maximum IOU is less than a threshold
$th_{iou}$, we mask out its value and exclude from the calculation
of backpropagation. Also, the suffix $k^{\prime}$ means the ground
truth whose bounding box is the best matched with $\hat{d_{b,k}}$.
The $b_{age}\left(d_{b,k^{\prime}}\right)$is also a boolean function
that masks out the value when ground truth has only bounding box annotation
and no age annotation. $L_{age\_single}\left(\hat{a},\,a\right)$is
a multi-task loss composed of mean variance loss \cite{pan2018mean}
and cross entropy loss,

\begin{align*}
L_{age\_single}\left(\hat{a},\,a\right) & =\lambda_{mean}L_{mean}\left(\hat{a},\,a\right)\\
 & +\lambda_{var}L_{var}\left(\hat{a},\,a\right)+\lambda_{ce}L_{ce}\left(\hat{a},\,a\right)
\end{align*}
The first two terms denote mean variance loss's components; mean loss
and variance loss. The last term denotes a cross entropy loss. The
$\hat{a}$ is a per year class vector which is the output of a softmax
activation. Finally, the summed up value is divided by $N_{a}$, which
denotes the number of faces NOT masked out. For gender estimation,
almost the same loss is used but binary cross entropy is used instead
of mean variance loss and cross entropy.

\begin{align*}
L_{gen} & =\frac{\lambda_{gen}}{N_{g}}\sum_{b}^{B}\sum_{k}^{K}b_{iou}\left(\hat{d_{b,k}},\,d_{b}\right)b_{gen}\left(d_{b,k^{\prime}}\right)\\
 & \times L_{bce}\left(\hat{g_{b,k}},\,g_{b,k^{\prime}}\right)
\end{align*}

\subsection{Training Strategy}

\subsubsection*{End-to-End Learning or Not}

Our proposed model is possible to train end-to-end with the loss function
above. The merit of end-to-end learning is that the intermediate feature
of the detection part can be optimized for age estimation. Meanwhile,
it might cause over-fitting to the training data. So, in the experiment,
freezing detection sub-network while training age estimation sub-network
is compared with end-to-end learning.

\subsubsection*{Tiling Augmentation}

\begin{figure}
\centering

\def\svgwidth{0.9\columnwidth}

\scriptsize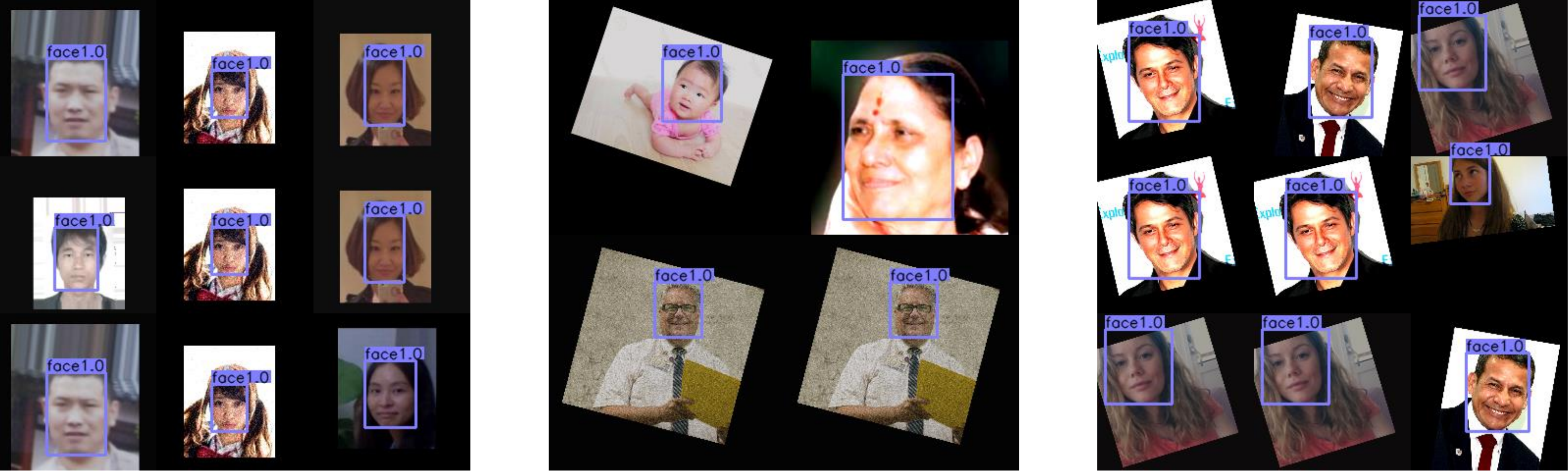

\caption{\label{fig:Two-samples-of}Two samples of tiling augmentation}
\end{figure}

It is desirable to train the model with multi-person photographed
datasets that have corresponding age ground truth but there are no
suitable datasets like that. Moreover, it is difficult to create as
mentioned above. Meanwhile, there are a lot of only face region annotated
datasets and only single face photographed datasets. So we introduce
tiling augmentation which utilizes the single person photographed
datasets for multiple person age estimation. In this method, in order
to create pseudo-multi-person photographed images, multiple images
are tiled patch-wisely as shown in Fig. \ref{fig:Two-samples-of}
with the number of images to use randomized. However, only feeding
the tiled images might mislead the model. It might learn there is
a person per patch. So we also feed the images from multiple face
detection dataset which has no age labels but has seamless images.
If these images are fed, the backpropagation for them in the age estimation
sub-network is ignored by the boolean masking function $b_{age}\left(d_{b,k^{\prime}}\right)$. 

\section{Experiments}

Our experiments roughly consist of two parts. One is the abrasion
study on multi-person age estimation. The other is abrasion study
and benchmarking on several widely used datasets that contain only
one person is photographed per image.

\subsection{Datasets}

We studied our method on three datasets: Images of Groups (IoG) \cite{gallagher2009understanding},
IMDB-WIKI \cite{Rothe-ICCVW-2015}, UTKFace \cite{zhang2017age} utilizing
the other three datasets: CLAP2016 \cite{escalera2016chalearn}, Mega
Age Asian \cite{zhang2017quantifying} and WiderFace \cite{yang2016wider}.
We chose IoG because it is the only dataset openly available with
multi-person age labels. Others were chosen because they release pre-cropped
images or they have plenty of images.

\textbf{Images of Groups} consists of 5080 group images where 28,231
faces were annotated with age group labels and gender labels \cite{gallagher2009understanding}.
To the best of our knowledge, this is the only dataset releasing multi-person
photographed images annotated with their ages. This dataset is challenging
because face sizes are small. The median distance between the eye
centers is 18.5 pixels. The definition of the age groups were 7 categories;
0-2, 3-7, 8-12, 13-19, 20-36, 37-65 and 66+. Because it is rough to
estimate coarsely, we used this dataset as an evaluation dataset and
didn't train with this dataset. Also, different from previous works,
we used a pre-cropped version of images.

\textbf{IMDB-WIKI} is the largest dataset with age and gender ground
truth crawled from IMDB (460,723 images) and WIKI (62,328 images)
\cite{Rothe-ICCVW-2015}. The age labels are distributed from 0 to
100. Although many of the previous works \cite{Rothe-ICCVW-2015,chen2017using,pan2018mean}
regard this dataset was not suitable for evaluation due to the noise,
we compared the evaluation result with other reported results since
it is one of the rare datasets which also provide pre-cropped images.
Following \cite{zhang2019c3ae}, We randomly split 20 \% of the data
as test data and remaining as training data. By the data cleaning
procedure explained bellow, we finally used 113,466 images as training
data.

\textbf{UTKFace} is also one of the rare datasets that offer pre-cropped
images with its age, gender, and ethnicity labels \cite{zhang2017age}.
It contains over 20,000 images and the age labels are per year level
from 0 to 116. we trained and evaluated with this dataset. We split
20\% of the data as test data similar to previous methods \cite{berg2020deep,cao2019rank,gustafsson2019dctd},
but we used from people of 0 to 100 years old as training data nonetheless
they trained and evaluated with the people of 21 to 60 years old.
However, we tested 21 to 60 years old people to compare with these
methods. By the data cleaning procedure, our training data was decreased
from 19,285 to 18,396 images.

\textbf{CLAP2016} is the dataset which contains 7,591 images with
corresponding age labels from 0 to 100 \cite{escalera2016chalearn}.
Their images have somewhat wide mergin arround faces. Because they
also provide apparent age labels and most of the previous works compare
with the apparent age estimation task, we don't evaluate on this dataset
and only used for training. By the data cleaning procedure, our training
data was decreased from 5,613 to 4,927 images.

\textbf{Mega Age Asian} contains 40,000 cropped facial images with
age labels \cite{zhang2017quantifying}. The age distribution is from
1 to 69. We used this dataset because the datasets above contain less
Asian faces. This dataset was used only at the training phase. By
the data cleaning procedure, our training data was decreased from
40,000 to 36,687 images.

\textbf{WiderFace} is a challenging dataset for face detection that
contains wide variability of images in scale, pose, occlusion, expression,
makeup, and illumination \cite{yang2016wider}. 32,203 images and
393,703 faces are included. We utilized this dataset to train a face
detector which was used to discard some noisy training images in the
age estimation dataset and create the ground truth of facial bounding
boxes. Also, this dataset was used in the tiling augmentation method.

\subsection{Implementation Details}

\subsubsection*{Data cleaning and Annotation}

As described in section 2.1, we used CenterNet \cite{zhou2019objects}
whose backbone is DLA-34 \cite{yu2018deep} as the detection sub-network
and as the annotator of facial bounding boxes. The CenterNet was trained
with the training set of WiderFace dataset. Even if there are official
annotations of bounding boxes in age estimation datasets, we trained
the all models with the definition of WiderFace's facial bounding
boxes. Many of the age estimation datasets especially the pre-cropped
version of datasets sometimes contains not annotated faces in the
background unconsciously and they can disturb the training of the
multi-face age estimation model. So, At the annotation time of bounding
boxes, we excluded images in which multiple faces are detected from
the training data except those of IoG dataset. On the other hand,
we didn't exclude from test data to compare the result with previous
methods. Instead, we regarded the largest detected face as the target
face and compared with ground truth age or gender.

\subsubsection*{Details of Model Architecture}

For the detection sub-network, images were input after resized into
$480\times480$. On the other hand, for the age estimation sub-network,
faces were cropped corresponding to the detections and resize into
$160\times224$ from high-resolution images. The cropped regions have
a 20 \% mergin from the detected bounding boxes at the upper boundary
and a 10 \% margin at the other boundaries to be robust to the detection
error. In all experiments, the CenterNet is used as the detection
sub-network and trained from the pre-trained weight created in the
above procedure. 

It was reported that VGG \cite{simonyan2014very} could estimate more
accurately than ResNet \cite{he2016deep} on the age estimation task
\cite{mallouh2019utilizing} and many previous works used VGG-16 as
a backbone for age estimation model \cite{pan2018mean,Rothe-ICCVW-2015,agustsson2017anchored}.
However, VGG-16 is around 4 times slower than ResNet34 and needs 4.5
times larger memory \cite{canziani2016analysis}. These days, TResNet
is proposed which is a little faster at inference time and a little
more accurate than ResNet \cite{ridnik2020tresnet}. Because multi-person
age estimation in real-time needs a faster and memory-efficient model,
we used TResNet instead of VGG-16. Also, we newly defined TResNet-S
which had the almost same parameters with ResNet34 by decreasing the
number of residual blocks from $(3,\,4,\,11,\,3)$ to $(3,\,3,\,7,\,3)$
per stage and channel width factor from 1.0 to 0.9 of TResNet-M, which
has about the same parameters with ResNet50. In the case the intermediate
feature connection was used, the original DCN-34's output feature
before the additional skip connection added for CenterNet was branched
off. The feature was 1/4 of the input size and the channel width was
48. Then, it was affine transformed into $10\times14$ size per the
region of bounding box prediction. Then the feature was concatenated
with the output of the residual block in age estimation model whose
output firstly became $10\times14$. To prove the performance improvement
was not stem from increased parameters, we reduced the channel width
of this residual block by 48. This results in a little light weight
model. The output of the age estimation backbone was branched into
two passes for age distribution prediction and gender prediction.
Each pass consisted of a $1\times1$ convolution layer, a global average
pooling layer, a dropout layer whose dropping rate was 0.2, and a
101-way or a 2-way fully-connected layer with a softmax activation.

\subsubsection*{Train and Test Settings}

For IMDB-WIKI dataset, we trained the detection sub-network from pre-trained
weight and the age estimation sub-network from scratch with adam optimizer
and a batch-size of 24 for 50 epochs, while the learning rate started
from 1e-4 and multiplied 0.1 at 30 and 40 epochs respectively. For
other datasets, we trained the model from the weight trained with
IMDB-WIKI with a batch-size of 12 for 80 epochs, while the learning
rate started from 1e-4 and multiplied 0.1 at 50 and 70 epochs respectively.
We applied random flip, random scaling, random crop, random rotation,
color jittering, random blur as default augmentation. When tiling
augmentation was applied, the number of tiles per image was randomly
chosen from 1, 4, or 9. For the sake of time efficiency, maximum of
four images were loaded and some tiles contained the same images.
In all experiments bellow, hyperparameters for loss function $\{\lambda_{size},\,\lambda_{off},\,\lambda_{mean},\,\lambda_{var},\,\lambda_{ce},\,\lambda_{gen}\}$
were set as $\{0.1,\,1.0,\,0.01,\,0.0025,\,0.05,\,0.1\}$ and $th_{iou}$
was set as 0.3. At the test time, predictions whose confidences were
greater than 0.2 were regarded as faces. In the training time, the
parameter $K$, the maximum processable number of faces, was set as
9 when tiling augmentation was applied, in other cases, set as 1 to
train fast. When multiple datasets were used, we defined one epoch
corresponding to the number of images in the biggest dataset, and
images were chosen randomly from each dataset with the weighted probabilities
based on the size of each dataset. The running time was tested on
Ubuntu 18.04, Intel Xeon pratinum 8259CL processor, a Tesla T4 GPU,
CUDA 10.0, python 3.6, and pytorch 1.4.

\subsection{Ablation Study on Multi-Person Age Estimation}

We tested on the test split of IoG dataset defined in \cite{gallagher2009understanding},
which contained around 20\% of faces. Because the test split is based
on the already cropped face images, there are mixtures of people to
evaluate or not to evaluate on the same pre-cropped images. So, we
estimated all faces but evaluated with only the faces for test set.
The test set was balanced among age groups and gender. In these experiments,
we trained the model with the training set of UTKFace, CLAP2016, Mega
Age Asian, and supporting detection dataset of WiderFace from the
pre-trained weight on IMDB-WIKI. At the test time, the parameter K,
the maximum processable number of faces, is set as 20 different from
the training time. 

The experimental results were shown in Table \ref{tab:Ablation-study-on}.
The \textquotedblleft baseline\textquotedblright{} was almost the
same model with Mean Variance Loss \cite{pan2018mean} except using
a light weight backbone. The \textquotedblleft intermediate feature\textquotedblright{}
denotes our proposed model using the intermediate feature connection
aimed to consider surrounding pixels to some extent. It improved age
estimation accuracy slightly despite a bit smaller model parameters
but gender estimation did not improved. The tile augmentation further
improved accuracy. Finally \textquotedblleft end-to-end\textquotedblright{}
denotes it was trained end-to-end meanwhile all of the above settings
trained with the detection sub-network frozen. The result shows that
end-to-end learning can be possible but a bit deteriorates the performance.
It might be because the detection pert was also tried to estimate
age and it caused overfitting. In our environment, the inference speed
of the ``baseline'' model was 14.3 fps and that of our proposed
model was 14.6 fps if the maximum detectable number $K$ was 20. When
$K$ was set as 1, 3, 10, and 20, the inference speed were 24.3, 23.1,
18.4, and 14.6 fps respectively. Because people don't move so fast,
these speeds are enough for the usual real-time analysis.

From this experiment, It can be said that detecting and estimating
with a single model enhance the accuracy for multi-person age estimation
together with tiling augmentation. 

\begin{figure*}
\centering

\def\svgwidth{2.0\columnwidth}

\scriptsize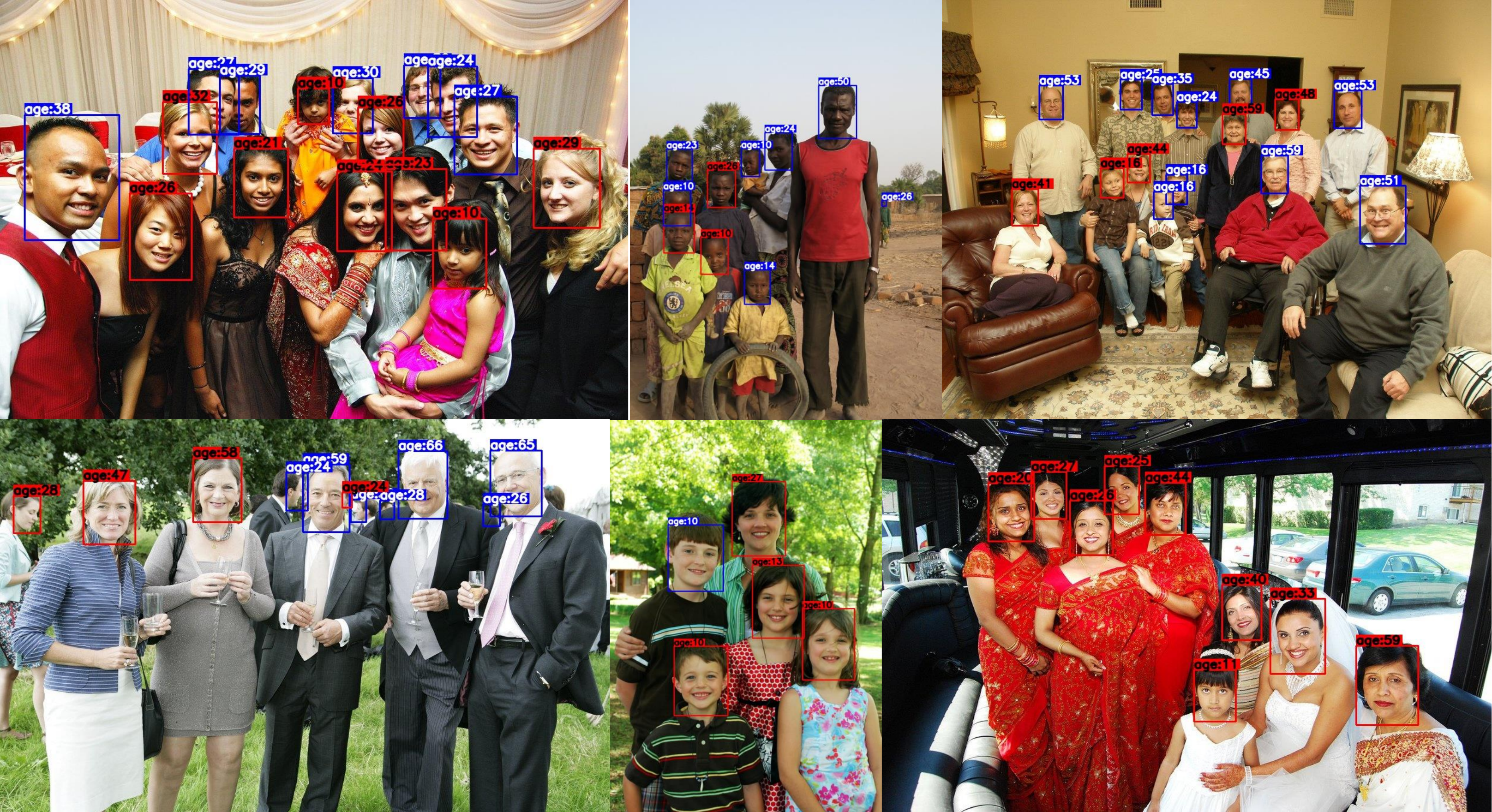

\caption{\label{fig:Visualization-of-the}Visualization of the result on IoG
dataset.}
\end{figure*}

\begin{table}
\caption{\label{tab:Ablation-study-on}Ablation study on IoG dataset.}

\noindent\resizebox{\columnwidth}{!}{%

\begin{tabular}{cccc|ccc}
 & intermediate & \multirow{2}{*}{tile aug} & \multirow{2}{*}{end-to-end} & \multicolumn{3}{c}{accuracy {[}\%{]}}\tabularnewline
 & feature &  &  & age & 1-off{*} & gender\tabularnewline
\hline 
\hline 
baseline &  &  &  & 38.78 & 79.04 & \textbf{88.59}\tabularnewline
\multirow{3}{*}{ours} & $\checkmark$ &  &  & 38.87 & 79.30 & 86.83\tabularnewline
 & $\checkmark$ & $\checkmark$ &  & \textbf{39.71} & \textbf{81.07} & 87.50\tabularnewline
 & $\checkmark$ & $\checkmark$ & $\checkmark$ & 38.45 & 80.10 & 87.35\tabularnewline
\end{tabular}

}

{\small{}{*}1-off accuracy of age group estimation}{\small\par}
\end{table}
By the way, all of the age accuracy even from the baseline model which
was almost the same as the state of the art method \cite{pan2018mean}
was low. We investigated the reason and it turned out that very young
value and very old value can't be output as shown in Table \ref{tab:Detail-result-on}
and Fig. \ref{fig:Visualization-of-the}. The reason might be derived
from biased distribution of training data and the mean averaging way
of age output that tend to be affected by far away age classes. 
\begin{table}
\caption{\label{tab:Detail-result-on}Detail result on IoG dataset. The result
using intermediate feature connection and tile augmentation is shown.}

\noindent\resizebox{\columnwidth}{!}{%

\begin{tabular}{cccccccc}
real age & 0-2 & 3-7 & 8-12 & 13-19 & 20-36 & 37-65 & 66+\tabularnewline
\hline 
\hline 
age & 0.0 & 0.0 & 55.3 & 28.0 & 72.3 & 66.9 & 54.4\tabularnewline
1-off & 0.0 & 86.6 & 88.7 & 97.0 & 99.3 & 99.3 & 98.7\tabularnewline
gender & 79.6 & 84.4 & 82.0 & 87.3 & 97.3 & 98.0 & 98.7\tabularnewline
\end{tabular}

}
\end{table}

\subsection{Ablation Study on Single Person Age Estimation}

Although our method is aimed to estimate multi-person, the conceptual
architecture of \textquotedblleft focusing on objects and still seeing
surroundings\textquotedblright{} may be effective even if only a single
face is shown in an image. So we also tested with single person photographed
datasets. Firstly we conducted an ablation study again on IMDB-WIKI
dataset. The results were Table \ref{tab:Ablation-study-on-1}.
\begin{table}[h]
\caption{\label{tab:Ablation-study-on-1}Ablation study on IMDB-WIKI dataset.}

%\noindent\resizebox{\columnwidth}{!}{%

\begin{tabular}{ccccc}
 & intermediate & \multirow{2}{*}{end-to-end} & age & gender\tabularnewline
 & feature &  & MAE & accuracy\tabularnewline
\hline 
\hline 
baseline &  &  & 5.87 & 92.23\tabularnewline
\multirow{2}{*}{ours} & $\checkmark$ &  & 5.82 & 92.18\tabularnewline
 & $\checkmark$ & $\checkmark$ & \textbf{5.64} & \textbf{92.82}\tabularnewline
\end{tabular}

%}

{\small{}{*}1-off accuracy of age group estimation}{\small\par}
\end{table}

\begin{table}[b]
\caption{\label{tab:Comparison-with-the}Comparison with the state of the art
methods on IMDB-WIKI.}

\begin{tabular}{cccc}
 & backbone & param.{*} & MAE\tabularnewline
\hline 
\hline 
C3AE \cite{zhang2019c3ae} & SSR \cite{yang2018ssr} & 0.04M & 6.91\tabularnewline
C3AE \cite{zhang2019c3ae} & C3AE & 0.04M & 6.48\tabularnewline
\cite{liu2018smart} & multi-model & - & 5.93\tabularnewline
\cite{liu2018age} & Xception \cite{chollet2017xception} & 22.9M & 6.93\tabularnewline
\cite{liu2018age} & multi-model & - & 5.89\tabularnewline
ours & TResNet-S & 19.2M & \textbf{5.64}\tabularnewline
\end{tabular}

{\small{}{*} The numbers of the parameters in the age esitmation backbones
are listed.}{\small\par}

%%%different split
%%%[Deep Modeling of Human Age Guesses for Apparent Age Estimation]	5.27 28.7M
%%%[Age and Gender Prediction from Face Images Using Convolutional Neural Network] 7.22 11.0M

\end{table}

Even for a single person situation, our proposed method enhanced the
accuracy in spite of a bit small model size. In this time, end-to-end
learning enhanced the accuracy. It might be because IMDB-WIKI has
a wide variety of images and hence overfitting did not occur. Our
results on IMDB-WIKI were compared with the sate of the art methods
in Table \ref{tab:Comparison-with-the-1} and outperformed their accuracy.
Although our model is design to be faster than usual methods, C3AE
\cite{zhang2019c3ae} used much smaller model. 

\begin{table}
\caption{\label{tab:Comparison-with-the-1}Comparison with the state of the
art methods on UTKFace with 21 to 60 years old people. Our method
was trained with 0 to 100 years old people different from others.}

\begin{tabular}{cccc}
 & backbone & param. & MAE\tabularnewline
\hline 
\hline 
OR-CNN \cite{niu2016ordinal}{\dag} & ResNet34 & 21.8 & 5.74\tabularnewline
CORAL \cite{cao2019rank} & ResNet34 & 21.8 & 5.39\tabularnewline
DCDT \cite{gustafsson2020energy} & ResNet50 & 25.6 & 4.65\tabularnewline
RandomizedBins \cite{berg2020deep} & ResNet50 & 25.6 & 4.55\tabularnewline
ours & TResNet-S & 19.2 & 4.61\tabularnewline
ours{*} & TResNet-S & 19.2 & \textbf{4.49}\tabularnewline
\end{tabular}

{\small{}{\dag} The result was reported in \cite{cao2019rank}.}{\small\par}

{\small{}{*} The model trained for IoG dataset in section 4.3 was
used.}{\small\par}

%%%different split
%%%4.87 age estimation using specific domain transfer learning ##vgg16
%%%7.21 age estimation system using deep residual netowrk classificaiton method##resnext 50
\end{table}

We also conducted an experiment on UTKFace dataset. The best method
from the ablation study which used the intermediate feature connection
with end-to-end learning was used. Our experiment was challenging
because previous methods trained and tested with from 21 to 60 years
old people but we trained from 0 to 100 years old people. Furthermore,
only our method used pre-cropped images. This means some faces were
wrongly cropped although previous methods used correctly cropped images
with the help of human checks. The result is Table \ref{tab:Comparison-with-the}.
You can see our method achieved a competitive result. Moreover, the
model trained with CLAP2016 and Mega Age Asian for IoG dataset outperformed
state of the art. It is interesting because in this experiment, more
than half of the faces are Asian faces but UTKFace contains mostly
Western faces. It can be said that even different race of faces is
helpful to estimate.

\section{Conclusion}

Different from most of the previous methods, we aimed to enhance the
accuracy of age estimation on multi-person photographed images. We
proposed a multi-person age estimation method to cope with facial
detection and age estimation in a single model. Thanks to the concept
of focusing on faces and still seeing surroundings realized by facial
cropping connection and intermediate feature connection, our method
further improved from the baseline model which utilized state of the
art method \cite{pan2018mean}. Moreover, our concept of focusing
on faces and still seeing surroundings was also useful to single person
photographed images and outperformed state of the art methods on IMDB-WIKI
dataset and UTKFace dataset. In the future, we will examine the effect
of our method on other tasks like human pose estimation with top-down
approaches which also need to detect and estimate from cropped regions.

{\small{}\bibliographystyle{ieee_fullname}
\bibliography{egbib}
 }{\small\par}
\end{document}